%% file: main.tex
\documentclass[conference,10pt]{IEEEtran}
\usepackage{xcolor}
\usepackage{amsmath}
\usepackage{url}
\usepackage{graphicx}

\begin{document}
%
\title{LEARNING STRATEGIES FOR SUCCESSFUL CROWD NAVIGATION}

\author{
    \IEEEauthorblockN{Rajshree Daulatabad*}
    \IEEEauthorblockA{York University \\
    Electrical Engineering \\ and Computer Science \\
    Email: rajshree@eecs.yorku.ca}
    \and
    \IEEEauthorblockN{Serena Nath*}
    \IEEEauthorblockA{York University \\
    Electrical Engineering \\ and Computer Science \\
    Email: serenath@eecs.yorku.ca}
}

\maketitle
\def\thefootnote{*}\footnotetext{Authors contributed equally to this work}\def\thefootnote{\arabic{footnote}}

\begin{abstract}
Teaching autonomous mobile robots to successfully navigate human crowds is a challenging task. Not only does it require planning, but it requires maintaining social norms which may differ from one context to another.  Here we focus on crowd navigation, using a neural network to learn specific strategies in-situ with a robot.  This allows us to take into account human behavior and reactions toward a real robot as well as learn strategies that are specific to various scenarios in that context.  A CNN takes a top-down image of the scene as input and outputs the next action for the robot to take in terms of speed and angle. Here we present the method, experimental results, and quantitatively evaluate our approach.
\end{abstract}
\IEEEpeerreviewmaketitle

\input{_1_Introduction/main.tex}
\input{_3_Method/main.tex}
\input{_4_Data_Collection/main.tex}
\input{_5_Learning_Model/main.tex}
\input{_6_Result/main.tex}
\input{_7_Extensions_and_Future_Work/main.tex}

\bibliographystyle{plain}
\bibliography{bib}

\end{document}

%% file: _1_Introduction/main.tex
\section{Introduction}

Robotics is rapidly advancing, with applications moving beyond the confines of the assembly line onto pedestrian walkways. This is made possible by advances in sensor and actuator technologies, significant increases in memory capacity and processing speed, digitized and shared information, easily accessible and programmable units, all at a reduced cost \cite{omarah2016}.  Robot applications such as security guards, retail floor assistants, transport vehicles inside airports and delivery assistants will require navigating indoor hallways, outdoor sidewalks and roads.  These applications will need to know how to navigate around humans, both individuals and crowds, in a socially acceptable manner.

Path planning and obstacle avoidance capabilities are insufficient to equip a robot with the ability to navigate crowds in a non-disruptive, socially acceptable manner. For example, Pradhan et al. \cite{pradhan2011robot} experimented on crowd navigation using dynamic path planning without capturing and learning human behaviour. Some methods manage pedestrians as moving obstacles, which may result in jerky, unnatural movements \cite{DBLP:journals/corr/ChenELH17}. Other models try to create more smooth motion by separating the problem into a prediction step (a set of predicted paths for pedestrians), and a planning step (choosing a safe path).  This may lead to the freezing robot problem where the available space is entirely occupied by predicted paths \cite{trautman2015robot}. In reality there is an interaction between the main agent and surrounding pedestrians that allows for each to navigate the space cooperatively. 

Many have attempted to create autonomous navigation systems that exhibit pro-social crowd navigation, using either model-based or learning-based approaches.  On the face of it, a model-based approach seems reasonable.  Many use manually designed models based on reasonable assumptions about the rules governing human behavior in crowds, or models derived from cognitive and social psychology \cite{luber2012socially}.  Warren develops a pedestrian model of crowd behavior, where collective motion emerges from basic behaviors such as steering, obstacle avoidance, target interception and pedestrian interactions \cite{warren2018collective}.  Other approaches attempt to mathematically model crowd behavior using principles of fluid dynamics, specifically particle flow, assuming that pedestrians self-organize and their strategies are selected simply based on crowd density.  Sparse crowds afford a similar pace and employ distance maintenance, while variations in speed and direction are required in more dense environments \cite{karamouzas2014universal}.

While many of these models demonstrate reasonable success, they do so within the domain of simulation.  It is questionable whether these would scale to the real world, given there is assumption upon assumption in modelling of the agent, the world and changes in the environment. Rather than building models based on a theoretical hypotheses about pro-social human behavior, with untested assumptions, it seems an approach that uses some sort of learning mechanism that operates in the real world deserves consideration.

Some of these learning-based approaches leverage data sets of human crowd behavior.  Luber et al. \cite{luber2012socially} use publicly available surveillance data to learn a set of dynamic motion prototypes, which are then used to compute dynamic cost maps for any-angle A*.  While using real-world data to do learning eliminates the need for some untested assumptions, there are key gaps. It is questionable whether a human navigating among humans is an adequate proxy for a robot.  In the long term robots may be ubiquitous, bipedal and indistinguishable from humans, but in the near term it is very likely that human reaction to robots is an essential component that needs to be addressed.

Another key drawback is that surveillance data is too coarse-grained, being a birds-eye view. It does not capture the incredible variety of fine-grained human interactions. These include overt communication such as gentle pressure and touch, tapping on the shoulder, natural language, eye contact, facial expression, hand/arm gestures. Furthermore, contextual cues may change the behavioral response of surrounding humans. People may evade rather than invade personal space \cite{Merrill.2012}, slowing down or speeding up based on the presence of crying children, someone staring at the ground, texting, being stuck behind a slow person or when there is a perceived threat to safety or comfort.  Furthermore, there is evidence of other demographic differences between humans that result in behaviors that indicate they belong to different populations in an otherwise homogeneous crowd \cite{henderson1972sexual}.  It’s unlikely surveillance video can capture this level of interaction.

The most significant drawback of learning crowd navigation from surveillance data is even if these human-to-human interaction behaviors were visible, it is very unlikely they would extend to human-to-robot interaction. It is not clear a human would tap a robot on the shoulder to communicate.  Even if the human did, it is unlikely the robot would be equipped with the right sensors to respond.  Conversely a robot tapping a human on the shoulder may result in a very different response than a tap from a human. It seems reasonable that successful crowd navigation will require learning that leverages actual robot-human interaction data.

Recently there has been much research that leverages statistical machine learning to learn crowd navigation.  Some use reinforcement learning to ensure social norms are followed.  Chen et al. \cite{DBLP:journals/corr/ChenELH17} note these methods are plagued by significant computational cost (calculating all possible paths of those nearby) as well as incomplete information in specifying norms (not taking into unobservable behavior such as pedestrian goals and ignoring the differences between individuals).  It is questionable how generalizable these are.  Instead they focus on learning what not to do, namely violations of social norms.  Unfortunately this method also requires significant training and does not take into account groups of people walking together.

We propose using a statistical machine learning approach to learn pro-social behaviors for navigating a university hallway.  To ensure the data captures robot-human interaction, and these interactions are pro-social, we train the robot based on observations while it is being tele-operated. To account for differences between pedestrians, we used multiple people to perform the tele-operation. To ensure we accounted for various scenarios that would typically occur we collected data before, during, and after class. To maximize our observations of crowd behavior, we collected data in close proximity to a major lecture hall and the main elevator.
Furthermore, instead of learning the whole trajectory in a dynamic environment, we chose to learn on a frame-by-frame basis. Not only does this reduce computational costs, but it also provides a focus on learning the pro-social behavior itself.  As noted above, trajectory planning is not the focus of our current work.


%% file: _3_Method/main.tex
\section{Method}
The components of our project included robot selection, identifying crowd scenarios, building a vision tracking system, collecting training data, building neural networks and training these, and finally testing and evaluating our overall model.
\subsection{Robot selection and configuration}
We used a Pioneer P3-DX mobile robot. Pioneer P3-DX has two sets of differential drive wheels and an aluminum body making it lightweight and rugged. Although the robot was equipped with various on board sensors, we did not use these.  We utilized a joystick interface connected via bluetooth to controlling the robot's navigation. 

The configuration of the Pioneer required both the joystick teleop configuration and bluetooth setup. We also needed to configure the ROS APIs, including ROSAria and the roscpp API.
The open source ARIA library was used as the ROS interface for the Pioneer. The RosAria node enabled communication with the robot's embedded controller.  It translated velocity, and acceleration commands to control the battery voltage and wheels \cite{wikiros_rosaira}.  Once the RosAria interface was setup, using the ROS Teleop node feature, the joystick was configured on the Pioneer via bluetooth. We mapped joystick buttons for forward, backward and angular movements. These controlled both the speed and movement of the robot.

\subsection{Crowd scenario identification}  Again as the focus of our project was learning crowd navigation strategies, the path planning aspect was kept very simple. The goal of the robot was to go from point A (one wall) to point B (opposite wall in the hallway), in a straight line. Very broadly, we identified three situations we wanted the robot to be able to navigate through:
\begin{enumerate}
\item When the path in front of the robot is clear
\item When the path in front of the robot has a few people 
\item When the path in front of the robot is occupied by a number of people, and groups of people
\end{enumerate}

\subsection{Building a vision-based tracking system} Our next objective was implementing a simplified variant of SLAM, using a neural network described under "Data collection" below. For crowd scenario identification some way of localization was needed. We used vision based tracking to identify the pedestrians and crowd, as well as the robot. A Kodak 4K camera was setup on tri-pod on the second floor of the Lassonde building, right above the first floor hallway area near the main administration office. The robot was marked with two different colors for identifying its location and orientation. As the robot was moving from a pre-defined point A to a pre-defined point B (and vice versa), our Kodak 4K wide-angle lens camera recorded video of the scene.

\subsection{Data collection for training} Approximately 4 hours of video recording was done over the period of 7 days.  Much of included trial and error, testing and prototyping. The video data used for the experiment and training was multiple trials going from point A to B and back for two hours taken on one day.
We wanted to cover a time that was typical for the semester, and considered various scenarios, including lunch time.  
Our trials spanned various scenarios including:
\begin{enumerate}
    \item Before the start of a class
    \item During a class
    \item At the end of the class
    \item During lunch
\end{enumerate}
The robot movement was controlled via joystick. The path was navigated for multiple trials. To ensure we captured variances between navigation, we used a few  team members during the trials. 

\subsection{Neural network training} 
The vision-based tracking system provides a data set containing 64x64 image of crowd labelled with robot orientation and details of the robot's next action, specifically speed and rotation in pixel units. This data is used an input to a second Convolutional Neural Network, CNN, which is trained to predict the next action for the robot. As the output actions are numeric values, a regression model was used. The neural network was trained based on the Mean Squared Error loss function. 

\subsection{Testing of the trained model} 
Ten percent of data collected was set aside for testing which was then used by the trained model for evaluation. The Mean Squared Error(MSE) was our parameter for testing. We chose Mean Squared Error over Mean Absolute Error as higher loss is penalized more than lower loss. The mean squared error is average of all squared residual for every data point \cite{reg_error}. Effectively, MSE describes the typical magnitude of the squared residuals.

The formal equation is shown below:
\newline

Mean squared error MSE = $\frac{1}{n}\sum_{t=1}^{n}e_t^2$ 

\subsection{Real world evaluation}
We planned to evaluate the trained model in the same environment and scenarios by feeding back the neural network input to navigate the robot. This could be done by manually inputting the commands into the joystick. To make this feasible we would do post-processing to discretize the continuous data (in 15 degree increments for example).  
Alternatively, we would close the loop by feeding the neural network output directly into the ROS joystick node and having it navigate the robot (with an override to provide for safety in case of imminent collisions).

We would then assess whether the robot successfully reached the goal based on two criteria:
\begin{enumerate}
\item The robot reaching the goal, from A to B, in a reasonable amount of time (determined by analyzing the training data trials)
    \item Whether chosen pedestrians, when surveyed, found the robot disruptive when compared to humans to their navigation goals.  
\end{enumerate}


%% file: _4_Data_Collection/main.tex
\section{Data Collection}
In order to learn behavioural strategies, robot motion and crowd information was needed in a simple manner. Visual data was captured from a top-down view of a hallway free of obstacles. Two points were chosen as goal location that the robot tried to reach. The robot was controlled by two separate operators for several runs each. Data was captured at multiple points throughout the day with high and low traffic.
\subsection{Camera calibration}
\begin{figure}
  \includegraphics[width=0.5\textwidth]{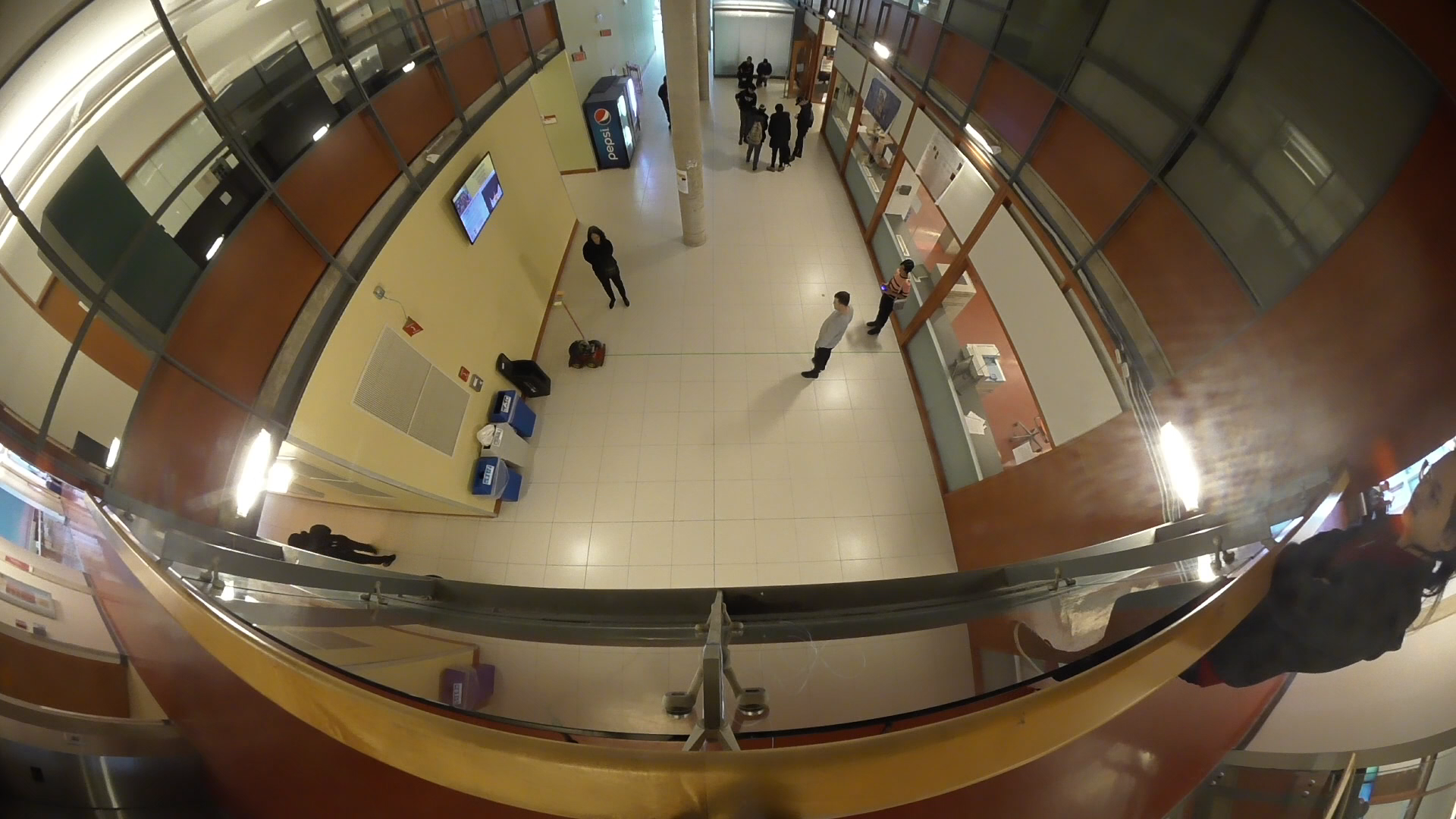}
  \caption{Raw camera footage prior to camera calibration.}
  \label{fig:raw}
\end{figure}
\begin{figure}
  \includegraphics[width=0.5\textwidth]{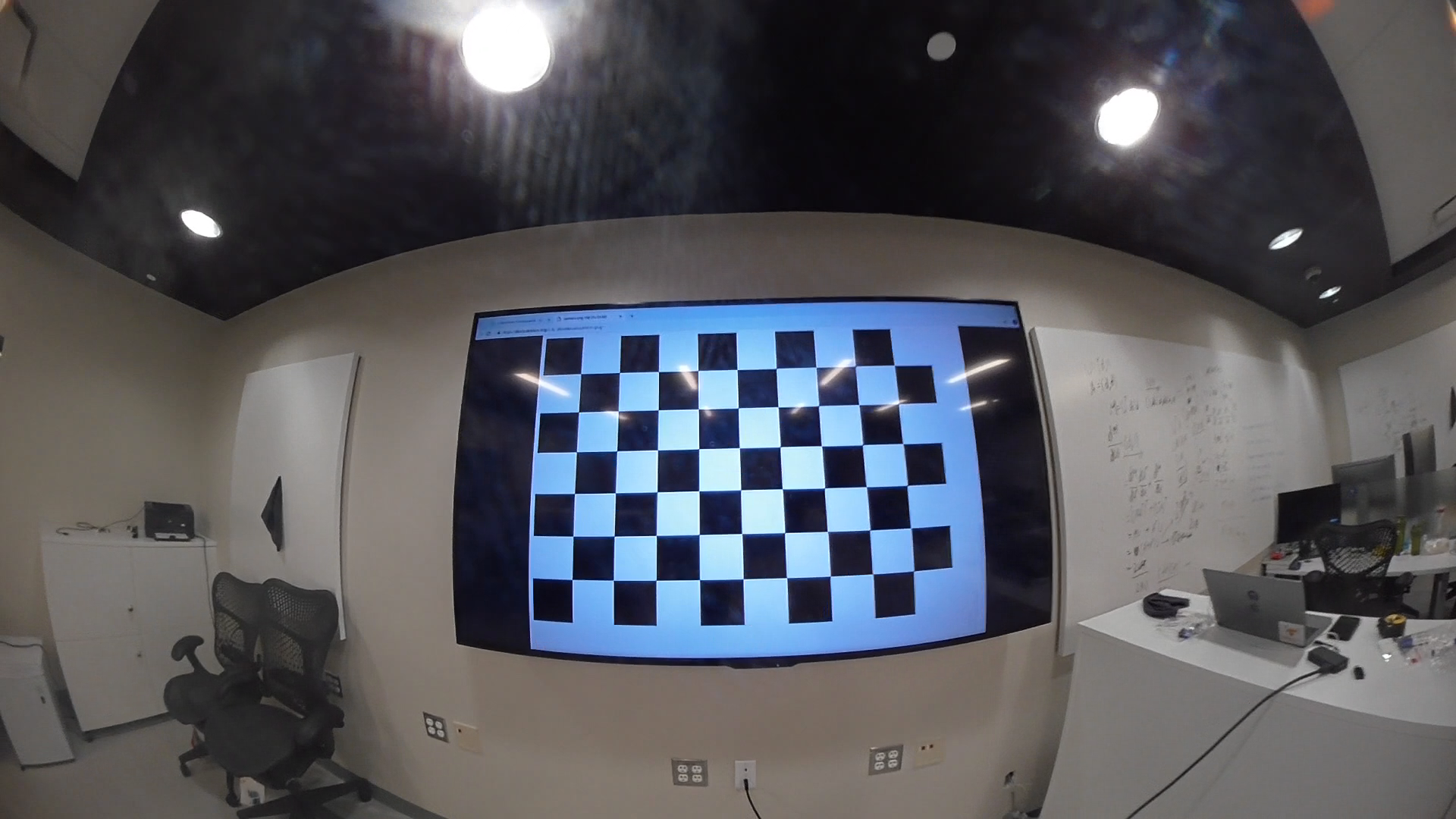}
  \caption{Checkerboard calibration card used to recover camera and distortion parameters.}
  \label{fig:cal}
\end{figure}
Visual data is captured with a Kodak PIXPRO SP360 4K which has an extremely large field of view. Standard camera calibration is performed by capturing views of a checkered black and white card with known dimensions as in Figure \ref{fig:cal}. Point correspondences made between the captured images of the calibration card are used to recover the camera intrinsics, 3 radial distortion coefficients, and 2 tangential distortion coefficients. The radial distortion process can be defined as:
\begin{align}
    \hat{x} &= x(1+k_1r^2+k_2r^4+k_3r^6) \\
    \hat{y} &= y(1+k_1r^2+k_2r^4+k_3r^6)
\end{align}
where $\hat{x}$, $\hat{y}$ are the distorted pixel coordinates, $k_1$, $k_2$, $k_3$ are the radial distortion coefficients, and $r$ is the distance of a point on the image from its optical center in pixels:
\begin{align}
    r^2 = x^2 + y^2
\end{align}
The tangential distortion process can be modeled as such:
\begin{align}
    \hat{x} &= x + [2p_1xy + p_2(r^2+2x^2)] \\
    \hat{y} &= y + [p_1(r^2+2y^2) + 2p_2xy]
\end{align}
where $p_1$, $p_2$ are the tangential distortion coefficients. Figure \ref{fig:raw} provides an example of an image before any processing.

\subsection{Homographic reprojection}
In order to simplify the localization of humans and the robot, the view must be transformed such that the floor in the view appears parallel with the camera view. This transformation is modeled as a homography:
\begin{align}
    \begin{bmatrix} 
    \hat{x}\\ \hat{y} \\ 1
    \end{bmatrix}
    =
    \begin{bmatrix} 
    h_1&h_2&h_3\\
    h_4&h_5&h_6\\
    h_7&h_8&h_9
    \end{bmatrix}
    \begin{bmatrix} 
    x\\ y \\ 1
    \end{bmatrix}
\end{align}
where $\hat{x}$, $\hat{y}$ are the pixel coordinates in the aligned image. Since the camera is stationary during data collection, the homography parameters are recovered once at the beginning of the footage. Four points on the original image are selected along with four corresponding target points at locations where the original points will move to after the homography. The parameters are recovered by minimizing this squared error loss with 80,000 iterations of gradient descent:
\begin{align}
    \mathcal{L}(\mathbf{x},\mathbf{y}) = \sum_{n=1}^N [(\hat{x_n}-x_n) + (\hat{y_n}-y_n)]^2
\end{align}
where $N$ is the number of points used. These values are computed using the following:
\begin{align}
    \begin{bmatrix} 
    \hat{x}\\ \hat{y}
    \end{bmatrix}
    =
    \frac{
        \begin{bmatrix} 
        h_1&h_2&h_3\\
        h_4&h_5&h_6
        \end{bmatrix}
        \begin{bmatrix} 
        x\\ y \\ 1
        \end{bmatrix}
    }{
        \begin{bmatrix} 
        h_7&h_8&h_9
        \end{bmatrix}
        \begin{bmatrix} 
        x\\ y \\ 1
        \end{bmatrix}
    }
\end{align}
The result of this reprojection can be seen in Figure \ref{fig:openpose}.

\subsection{Human and robot detection}
The highest point on the robot was marked with two orange and yellow markers perpendicular to the robot's front. The original plan was to track these coloured markers with simple colour matching. However, compression artifacts and lighting issues made that a difficult task to automate. The two markers were manually tracked by a human.

\begin{figure}
  \includegraphics[width=0.5\textwidth]{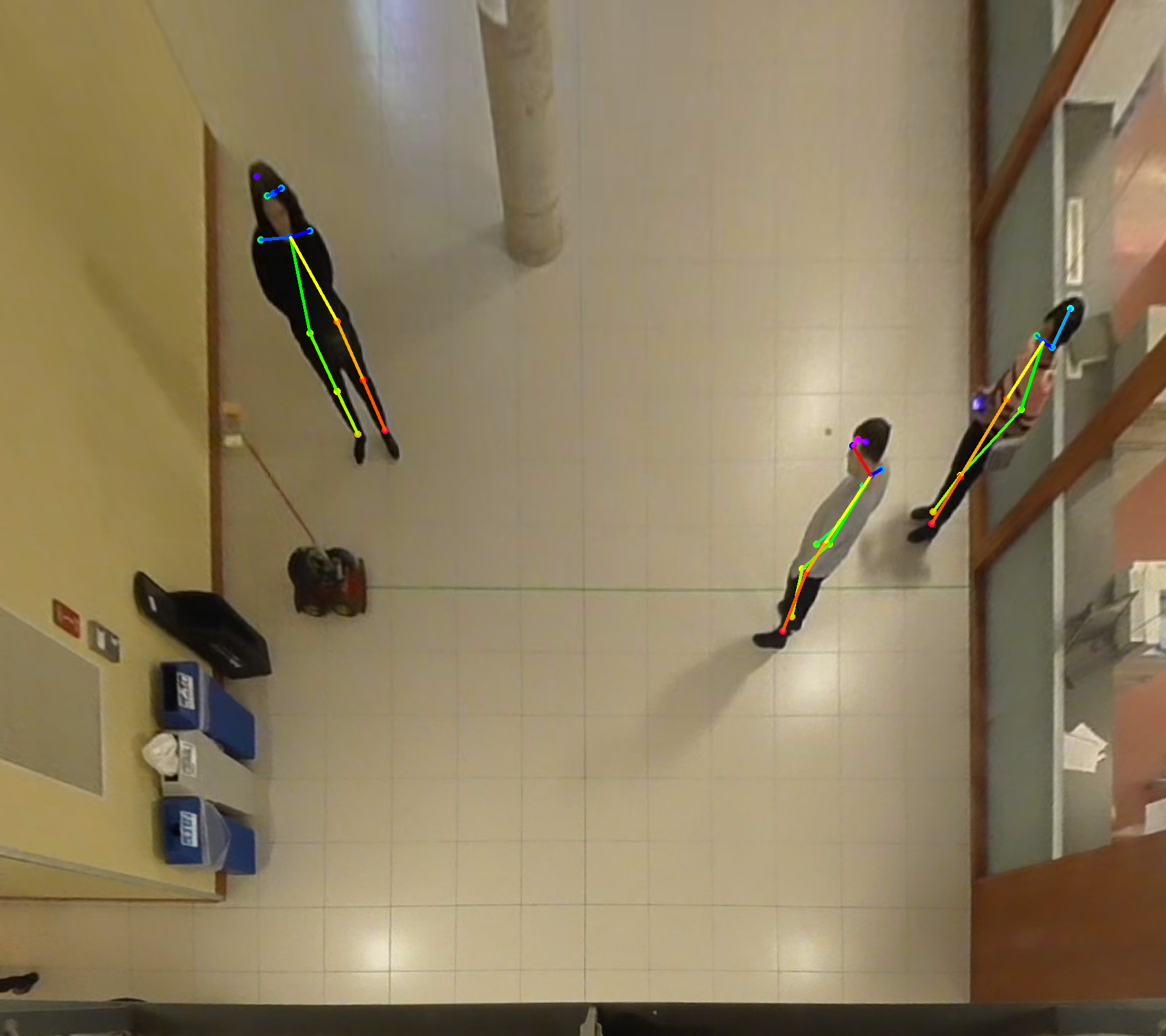}
  \caption{Three human subjects are detected with Openpose, highlighting their joints. The area is free of obstacles, and the image is aligned such that the floor is flat with the image. The robot is visible on the left near the trash bins.}
  \label{fig:openpose}
\end{figure}

Human tracking was done with \textit{Openpose} which does 2D human pose estimation. This pose estimation captures key joints in humans such as: left shoulder, right knee, neck, etc. For the purposes of this work, it was only necessary to find the position of humans on the floor; directly corresponding to their position in the world. The \textit{neck} key point is used as the position for every human detected. A visualization of these detections can be seen in Figure \ref{fig:openpose}.

\begin{figure}
  \includegraphics[width=0.5\textwidth]{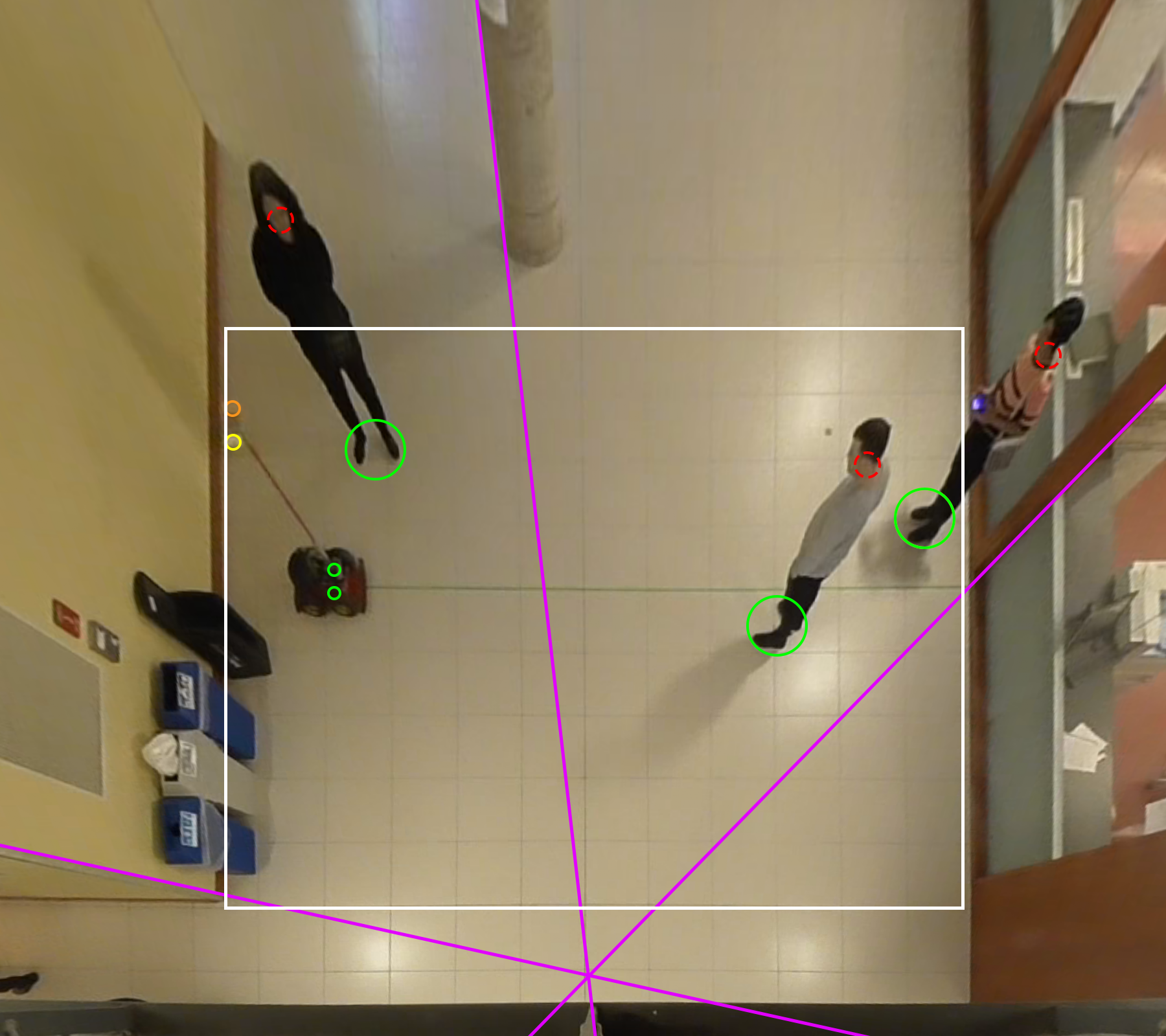}
  \caption{Purple lines represent continued lines from the world used to find the optical center. Their intersection is the optical center. The white square represents the unoccupied space that the robot may roam. Green circles are the corrected detections of humans and robot markers. Red dotted circles are the detected neck key points.}
  \label{fig:perspective}
\end{figure}
\subsection{Perspective correction}
Due to the image aligned relative to the floor, the height of the tracked targets needed to be accounted for to recover their positions on the floor. This can be done with the following:
\begin{align}
    x = \hat{x}(1-\frac{h_\text{target}}{h_\text{camera}}) \\
    y = \hat{y}(1-\frac{h_\text{target}}{h_\text{camera}})
\end{align}
where $x$, $y$ are the true positions on the floor, $\hat{x}$, $\hat{y}$ are the perceived position of the targets, $h_\text{target}$ is the height of the target above the floor, and $h_\text{camera}$ is the height of the camera above the floor.
The positions are relative to the point on the floor where the normal of the floor is inline with the camera's optical center.
This can be determined by finding the intersection of lines in the image that are normal to the floor in the world.
The height of the robot is known and constant but heights of humans can vary. However, this variance only introduces a small error and thus a constant height is assumed for all detected humans.

After correcting for perspective, the robot's position and orientation can be computed. Its position is the point between its two markers while its orientation can be computed also from the markers as they occupy the left and right side. A visualization of these corrections and lines used to find the optical center can be found in Figure \ref{fig:perspective}.

\subsection{Data normalization}
\begin{figure}
  \includegraphics[width=0.5\textwidth]{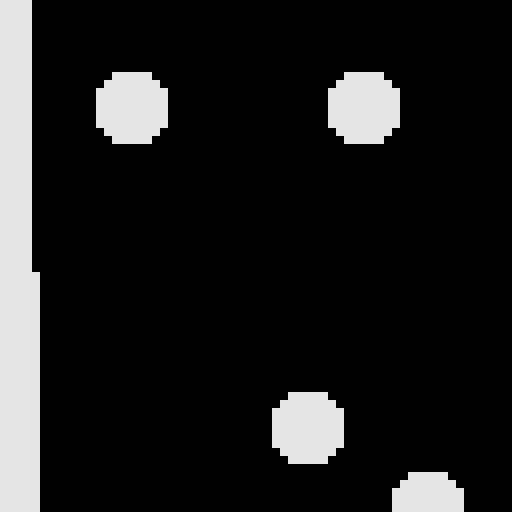}
  \caption{An instance of an occupancy map used in the data. White circles represent people while the white area on the left represents a wall. The map is relative such that the robot is always in the middle and is always trying to move to the right side.}
  \label{fig:data}
\end{figure}

To simplify the task for the neural network, the data is cast as a 64x64 occupancy map. This map is centered on the robot and is oriented relative to the target the robot is trying to reach. Orientation of the robot is also relative to this view. Humans are represented as circles that are occupied while the walls are sections of the map that are occupied as can be seen in Figure \ref{fig:data}.

%% file: _5_Learning_Model/main.tex
\section{Learning Model}
\begin{figure*}[!ht]
  \includegraphics[width=\textwidth]{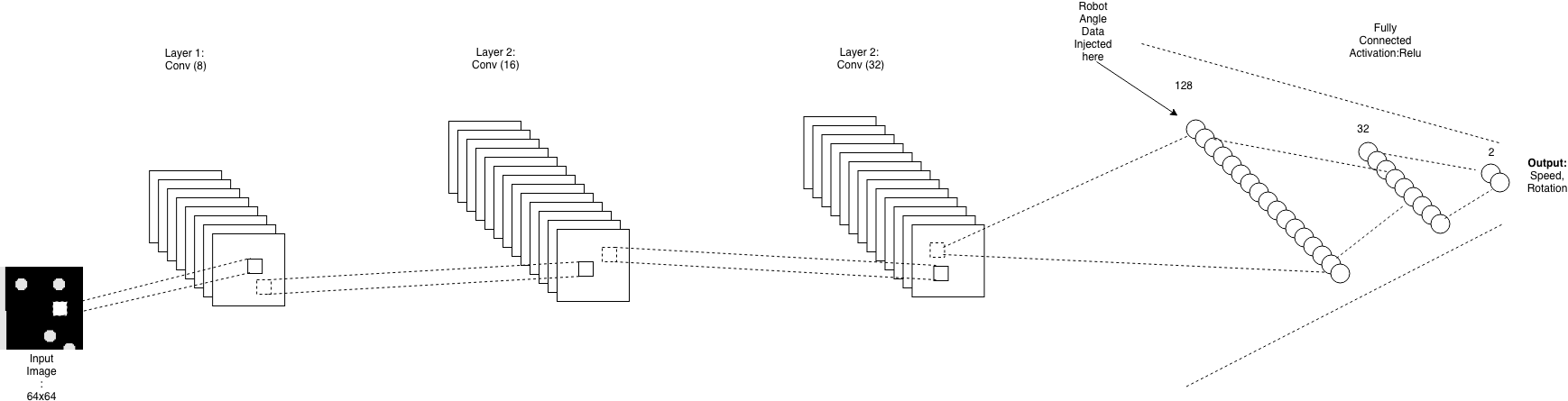}
  \caption{CNN architecture.}
  \label{fig:architecture}
\end{figure*}

Convolutional Neural Networks (ConvNets or CNNs) are one of the most utilized architectures for image recognition and classification, as typical neural networks do not scale well for processing full images \cite{medium_cnn}.  Three types of layers, convolutional, pooling and fully-connected, are typically used within the architecture.  CNNs usually apply the Softmax function to classify an object with probabilistic values between 0 and 1. 

In our case, instead of probabilistic output, we had numeric values as output denoting robot speed and angular movement. The CNN input consisted of robot orientation and pedestrian or crowd location (relative position to robot). The output was the new speed and angular rotation for the robot's next action.

\subsection{Input pre-processing}  
In the course of pre-processing the input, feature selection and feature normalization was performed. Feature selection is typically used for simplification of models, shorter training times, avoiding the 'curse of dimensionality', and enhancing generalization by reducing overfitting. Feature normalization, or scaling on the other hand, standardizes the range across all data data features to provide for better performance \cite{wiki_feature_scaling}. In our case, color in crowd identification was not needed. Instead we used grayscale values. Also, pixel values were converted to float with one value per pixel (grayscale). Pixel values were then divided by 255 to change the range to 0-1 as part of feature normalization.
       \begin{equation}
	x=x/255.0
\end{equation} 
Robot orientation data was converted to a 2D array of [sin(x),cosin(x)] where x is in the input robot orientation.
        \begin{equation}
	{x1,x2}=sin(\Theta),cos(\Theta)
\end{equation}
    
\subsection{Trial and error: fine-tuning the CNN Hyperparameters}
We obtained various sets of results based on variations of the neural networks we tried for learning.
Initially we setup a neural network for a well known data set, MNIST to ensure we had a working model.  The input was 28x28 pixels. This model showed convergence in less than an hour. We used an available laptop to do the training. We then modified this working model to take a 64x64 pixel input instead. We also modified the output for a regression rather than classification.  We limited this to speed only initially.  It showed no signs of learning (a reduction in the MSR) after 80 epochs and 3-5 hours of training.  We then adjusted various hyperparameters, including the learning rate, the number of filters, the batch size, and the types of optimization used.  We aborted various trials after 5 hours of no improvement in learning. We tried various permutations and combinations of hyperparameters over a two week period, running just under 30 sets (approximately 5 hours per trial, and 2 trials a day). 
We then switched to a more powerful GPU-accelerated machine (approximately 200 times faster).  We ran multiple trials adjusting hyperparameters. 
As seen in Figure \ref{fig:train_1}, we see a steady trend of decreasing loss until we reach a minimum. It seems the CNN successfully learned to provide an appropriate speed.
\subsection {Final CNN architecture} 
Following is the CNN architecture we finally settled on for training:
\begin{itemize}
    \item Convolution Layer 1. 8 5x5 Filters.
    \item Convolution Layer 2. 16 5x5 Filters.
    \item Convolution Layer 3. 32 5x5 Filters.
    \item Output of Convolution Layer 3 is flattened and the robot's orientation data is inserted.
    \item Dense layer 1 with 128 nodes. Activation:relu
    \item Dense layer 2 with 32 nodes. Activation:relu
    \item Dense layer 3 with 2 nodes. Activation:none
\end{itemize}
Three convolution layers with incremental filters followed by fully connected three hidden layers with ReLU activation functions \cite{stenroos2017object}. The convolutional layer applies an array of weights to all of the input sections from the image and creates the output feature map. Most CNNs have pooling layers. The pooling layers simplify the information that is found in the output from the convolutional layer. In our case we did not use a pooling layer. The last layer is the fully connected layer which effectively oversees the gathering of findings from previous layers and provides an N-dimensional vector, where N stands for the total number of classes or output regression values. Another way of reducing the data volume size is adjusting the stride parameter of the convolution operation. The stride parameter controls whether the convolution output is calculated for a neighbourhood centred on every pixel of the input image (stride 1) or for every nth pixel (stride n)\newline

\subsection{Training} 
The parameters we eventually decided upon for training this CNN included:

\begin{itemize}
\item Loss function = Mean Squared Error
\item Number of Epochs = 100,000
\item Batch Size = 128
\item Learning Rate = 0.0001 with decay e+0.00005
\end{itemize}

%% file: _6_Result/main.tex
\section{Results}
\begin{figure*}[!ht]
  \includegraphics[width=\textwidth]{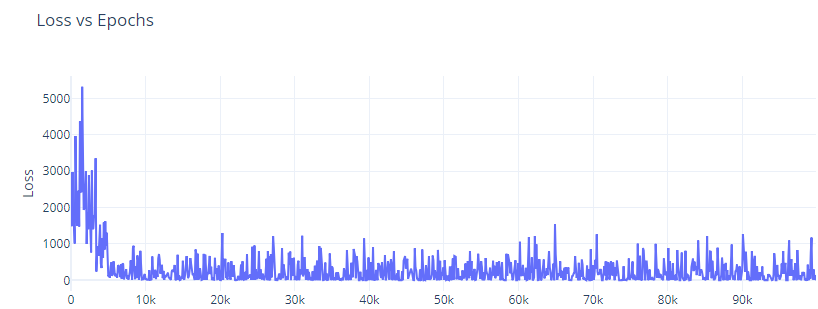}
  \caption{Loss vs. Epoch with multiple outputs}
  \label{fig:train_2}
\end{figure*}

\subsection{Final CNN performance}
\begin{figure}
\begin{centering}
  \includegraphics[width=0.5\textwidth]{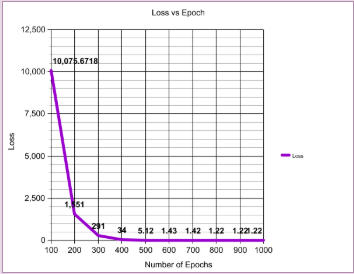}
  \caption{Loss vs. Epoch}
  \label{fig:train_1}
\end{centering}
\end{figure}

The CNN which was trained with only one output parameter, the robot's speed, successfully converged as shown in figure \ref{fig:train_1} and we were able to minimize the loss. Given the success of this initial CNN, we added a second output of rotational angle. Results of training this new model are in Figure \ref{fig:train_2}.  Here we see a decreasing trend of MSE, however there are a number of spikes. In total,  99,900 epochs were run. The result, in terms of average baseline speed (in cm/s) and rotation (in degrees) for training data and corresponding testing data, is as follows: 
\begin{itemize}
    \item Baseline: Speed average (cm/s): [22.617376], Rotation average (degrees): [13.952758]
    \item Using the Neural Net: Speed average (cm/s): [10.503133], Rotation average (degrees): [6.349384]
\end{itemize}

As we can see, the deviation in average speed and average rotation is 12 cm/s and 7 degrees respectively. This accounts for the spikes in the graph. We believe that the model would benefit from additional training data, and possibly improvements to the architecture and further hyper-parameter tuning. Unfortunately our data collection was cut short by a mechanical issue with the Pioneer which is still awaiting repairs.

In all, in this small scale experiment for crowd navigation we have demonstrated we are able to learn appropriate strategies for a robot in a university hallway.  
\subsection{Empirical real world testing of the trained model}
Unfortunately due to the mechanical issues we encountered during the end stages of data collection, we were not able to evaluate the robot in the environment. Our plan was to close the loop by controlling the robot with the joystick, based on the output from the trained neural network (post-processing to translate the speed and rotational angle into commands that could be fed back into the ROS joystick controller node).  As noted previously our evaluation criteria would have included both assessing whether the time the robot took to get from the destination to the goal fell within some range observed during the training trials, and surveying pedestrians to confirm the robot was not disruptive to them in any way, violating any obvious norms while navigating.    

%% file: _7_Extensions_and_Future_Work/main.tex
\section{Extensions and Future Work}
Extensions and future work include:
\begin{enumerate}
\item Completing the evaluation step mentioned above. Ideally by closing the loop and using the output to autonomously navigate the robot in the environment. 
    \item Adding additional sensors to the robot, including on-board cameras and auditory capabilities (allowing detection of noise volume, shouting, alarms, and perhaps natural language understanding). These additional inputs would provide for additional features to train upon. 
    \item Providing additional communication capabilities for the robot, including a face for facial expressions, including auditory - such as beeping or speech. 
    \item Running additional trials to collect more data, both in the real world and data augmentation.
\item Expanding to other related environments and scenarios beyond students going to class in a university hallway. 
\item Considering the orientation of each person in the crowd to provide for more successful behaviors. 
\item Training the same model to manage multiple other environments (such as the subway or hospital) and using the various cues (visual object identification or sound) to differentiate between these. 
 
\end{enumerate}

%% file: main.bbl
\begin{thebibliography}{10}

\bibitem{DBLP:journals/corr/ChenELH17}
Yu~Fan Chen, Michael Everett, Miao Liu, and Jonathan~P. How.
\newblock Socially aware motion planning with deep reinforcement learning.
\newblock {\em CoRR}, abs/1703.08862, 2017.

\bibitem{henderson1972sexual}
L.F. Henderson and D.J. Lyons.
\newblock Sexual differences in human crowd motion.
\newblock {\em Nature}, 240(5380):353, 1972.

\bibitem{karamouzas2014universal}
Ioannis Karamouzas, Brian Skinner, and Stephen~J Guy.
\newblock Universal power law governing pedestrian interactions.
\newblock {\em Physical review letters}, 113(23):238701, 2014.

\bibitem{luber2012socially}
Matthias Luber, Luciano Spinello, Jens Silva, and Kai~O Arras.
\newblock Socially-aware robot navigation: A learning approach.
\newblock {\em Intelligent robots and systems (IROS), 2012 IEEE}, 2012.

\bibitem{Merrill.2012}
Jamie Merrill.
\newblock How do i: Make my way through a crowd?
\newblock {\em Independent}, 2012-01-12.

\bibitem{omarah2016}
K.~O'Marah.
\newblock Robotics is coming faster than you think.
\newblock {\em Forbes}, 2016-08-18.

\bibitem{reg_error}
Christian Pascual.
\newblock {\em Understanding Regression Error Metrics}, (2018, Sep 28).
\newblock \url{https://www.dataquest.io/blog/understanding-regression-error-metrics/}.

\bibitem{pradhan2011robot}
Ninad Pradhan, Timothy Burg, and Stan Birchfield.
\newblock Robot crowd navigation using predictive position fields in the potential function framework.
\newblock In {\em American Control Conference (ACC), 2011}, pages 4628--4633. IEEE, 2011.

\bibitem{medium_cnn}
Prabhu Raghav.
\newblock {\em Understanding Convolutional Neural Networks (CNN), Deep Learning}, (2018, Mar 04).
\newblock \url{https://medium.com/@RaghavPrabhu/understanding-of-convolutional-neural-network-cnn-deep-learning}.

\bibitem{wikiros_rosaira}
Ros.org.
\newblock {\em ROSARIA Tutorial - ROS.org Wiki}, (2018, Aug 09).
\newblock \url{http://wiki.ros.org/ROSARIA}.

\bibitem{stenroos2017object}
Olavi Stenroos.
\newblock Object detection from images using convolutional neural networks.
\newblock Master's thesis, Aalto University, 2017.

\bibitem{trautman2015robot}
Pete Trautman, Jeremy Ma, Richard~M Murray, and Andreas Krause.
\newblock Robot navigation in dense human crowds: Statistical models and experimental studies of human--robot cooperation.
\newblock {\em The International Journal of Robotics Research}, 34(3):335--356, 2015.

\bibitem{warren2018collective}
William~H Warren.
\newblock Collective motion in human crowds.
\newblock {\em Current directions in psychological science}, 27(4):232--240, 2018.

\bibitem{wiki_feature_scaling}
Wikipedia.
\newblock {\em Wikipedia Page for Feature Scaling}, (2018, Nov 25).
\newblock \url{https://en.wikipedia.org/wiki/FeatureScaling}.

\end{thebibliography}
